\documentclass[twocolumn]{article}
\usepackage{usenix}
\usepackage{graphicx}
\usepackage{amsfonts}
\usepackage{amsmath}
\usepackage{epigraph}
\usepackage[utf8]{inputenc}
\usepackage{todo}

\title{Debugging Differential Privacy: A Case Study for Privacy Auditing} 

\author{Florian Tramèr\thanks{Authors ordered reverse alphabetically.}, Andreas Terzis, Thomas Steinke,  Shuang Song, Matthew Jagielski, Nicholas Carlini  \\ Google Research}
\date{}

\begin{document}

\maketitle

\iffalse
    \section*{Abstract}
    Backpropagation Clipping is a recently proposed technique to train neural network
    models with differential privacy.
    Inspired by recent advances in \emph{auditing} which have been used for estimating lower bounds on
    differentially private algorithms,
    here we show that auditing can also be used to find flaws in (claimed) differentially
    private schemes.
    We empirically audit Backpropagation Clipping and
    find, with $99.99999999\%$ confidence, that the
    implementation does not satisfy its differential privacy
    guarantees.
    %
    Our analysis shows that even papers with proofs of privacy
    should be audited in practice.

    \section{Introduction}
    
    \setlength\epigraphwidth{.3\textwidth}
    \epigraph{\emph{Beware of bugs in the above code; \\ I have only proved it correct, not tried it.}}{- Donald E. Knuth}
    
    A machine learning algorithm is
    differentially private 
    if an adversary cannot accurately distinguish between a model trained on
    dataset $D$ versus a model trained on a different dataset $D'$ differing in one 
    training example.
    More precisely, an algorithm is $(\varepsilon,\delta)$-differentially private  if any such distinguishing attack has a true positive rate at most $p(\varepsilon,\delta,q)$, dependent on the privacy parameters $\varepsilon$ and $\delta$ and the tolerable false positive rate $q$.
    Consequentially, we can argue that an algorithm is \emph{not}
    $(\varepsilon,\delta)$-differentially private by presenting an attack that accurately distinguishes two such input datasets.
    
    Backpropagation Clipping \cite{stevens2022backpropagation} is a recently proposed
    training mechanism that claims $(\varepsilon,\delta)$-differential privacy (DP).
    The authors' public implementation of Backpropagation Clipping
    trains a neural network on the MNIST dataset to $98.9\%$ accuracy with 
    a (proved) claim of $(0.21,10^{-5})$-DP.
    
    We show this privacy claim cannot be true.
    We construct MNIST$'$, an augmented version of the MNIST dataset
    that differs from MNIST by the addition of exactly one example.
    We then show that it is possible to distinguish 
    a model trained on MNIST from a model trained on MNIST$'$
    with false positive rate $q=0.174\%$ and true positive rate $4.922\%$, which is $22\times$ greater than the allowable true positive rate
    of $p(0.21, 10^{-5},0.174\%)=0.216\%$,
    refuting the paper's privacy claim.
    The authors acknowledged the correctness of our analysis,
    and agree their scheme is not private.
\else
    \section*{Abstract}
    Differential Privacy can provide \emph{provable} privacy guarantees for training data in machine learning.
    However, the presence of proofs does not preclude the presence of errors.
    Inspired by recent advances in \emph{auditing} which have been used for estimating lower bounds on
    differentially private algorithms,
    here we show that auditing can also be used to find flaws in (purportedly) differentially
    private schemes.
    
    
    In this case study, we audit a recent open source implementation of a differentially private deep learning algorithm and find, with $99.99999999\%$ confidence, that the implementation does not satisfy the claimed differential privacy guarantee.
    

    \section{Introduction}
    
    \setlength\epigraphwidth{.3\textwidth}
    \epigraph{\emph{Beware of bugs in the above code; \\ I have only proved it correct, not tried it.}}{- Donald E.~Knuth}
    
    A machine learning algorithm is
    differentially private 
    if an adversary cannot accurately distinguish between a model trained on
    dataset $D$ versus a model trained on a different dataset $D'$ differing in one 
    training example.
    More precisely, if an algorithm is 
    $(\varepsilon,\delta)$-differentially private \cite{dwork2014algorithmic} then 
    we can show that (loosely speaking) the ratio between
    true positive rate and false positive rate $TPR/FPR < e^\varepsilon$
    for any distinguishing attack (ignoring for now some small $\delta$-factor).

    A privacy audit applies this analysis in reverse:
    it constructs an attack that maximizes the $TPR/FPR$ ratio and thereby
    obtains an empirical lower bound on the privacy parameter $\varepsilon$.
    This has traditionally been used to assess the tightness of  differential privacy proofs~\cite{nasr2021adversary, jagielski2020auditing}.
    
    In this paper we show privacy audits can also 
    find bugs in differential privacy implementations,
    by showing that the measured lower bound exceeds the (claimed) upper bound.
    %
    Errors in implementations or proofs of differential privacy are surprisingly common, and can also be exceedingly hard to detect.
    Privacy auditing offers a way to easily detect some errors. We argue that it should become routine to perform such checks.
    To illustrate this point, we perform a real privacy audit and identify a significant bug in the implementation of a recent system.

    \paragraph{Case Study: Backpropagation Clipping}
    
     \cite{stevens2022backpropagation} is a recently proposed differentially private
    training mechanism accompanied by an open-source implementation.
    The system trains a neural network on the MNIST dataset to $98.9\%$ accuracy with a $(0.21,10^{-5})$-DP guarantee.

\begin{figure}
    \centering
    \vspace{-2em}
    \includegraphics[scale=.68]{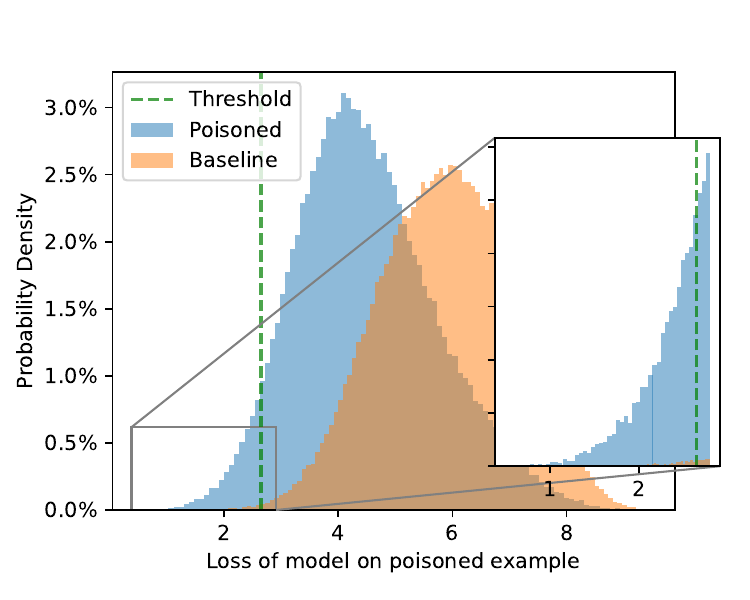}
    \vspace{-2em}
    \caption{The distribution of loss values 
    using a claimed $(0.21,10^{-5})$-DP algorithm.
    We train $100{,}000$ models on one dataset $D$ (in orange), 
    and another $100{,}000$ models on another $D' = D \cup \{x_p\}$
    (in blue).
    With a threshold $\tau=2.64$, our attack has a
    true positive rate of 4.922\% and false positive rate of 0.174\%.
    The Clopper-Pearson bounds allow us to
    show that $\varepsilon > 2.79$ with
    probability at least $1 - 10^{-10}$.
    This refutes the claim the algorithm is in fact $(0.21,10^{-5})$-DP\vspace{-.8em}}
    \label{fig:my_label}
\end{figure}

    This scheme is an excellent test case for privacy auditing as it has an easy-to-use and efficient open source implementation, is a real system (not contrived by us), and outperforms the state of the art by a factor of $30$. 
    
    Unfortunately, we find that the claimed differential privacy guarantee is not true.
    By training models on the MNIST dataset augmented with a single example, 
    in Figure~1 we show we can distinguish which dataset any given model was
    trained on
    with a true positive to false positive ratio that allows us to emperically
    establish a lower bound of $\varepsilon > 2.79$, a value $10\times$ higher
    than the claimed value of $\varepsilon=0.21$.
    An analysis of the implementation identifies a common category of bugs
    that we have found several times in the past.
    The authors graciously acknowledged the correctness of our analysis, and are
    fixing this issue by increasing the noise added to gradient descent in the implementation to match the theory.

\fi

\section{Background}

\paragraph{Machine learning notation.}
Let $f_\theta \gets \mathcal{T}(D)$ denote the neural network
model $f$ with learned parameters $\theta$, obtained by running
the (stochastic) training algorithm $\mathcal{T}$ on dataset $D$.
We consider labeled datasets $D = \{(x_i, y_i)\}_{i=1}^N$ 
with $N$ examples where each example $x_i$ (e.g., a picture of an
animal) is given a label $y_i$ (e.g., the label ``dog'').
Models are trained by minimizing the loss $\mathcal{L}(f_\theta, D)$
via stochastic gradient descent.

\paragraph{Private machine learning.}
When a model $f_\theta$ is trained on a dataset $D$, it is possible
that an adversary with access to $\theta$ might be able to learn
some information about the dataset $D$.
In the worst case, an adversary could
completely extract an individual training example $x \in D$ from the
dataset [cite].
While powerful, it can be difficult to reason about these attacks formally.
And so most work focuses on what is, in some sense, the ``minimal'' privacy: in a \emph{membership
inference} attack, an adversary only predict whether or not a model's
training dataset contains a particular
example $(\hat x, \hat y)$.
If it's not possible to detect between the presence or absence of
a particular example in a training dataset, then certainly it's not
possible to extract the example completely.
Therefore, preventing this weak attack also prevents any stronger attack.

Formally, membership inference attacks $\mathcal{A} : (f_\theta, (\hat x, \hat y)) \to \{0, 1\}$ take as input a trained model $f_\theta \gets \mathcal{T}(D)$, along with a query $(\hat x, \hat y)$ and output a
binary prediction of whether or not $(\hat x, \hat y) \in D$.
For a membership inference attack, the false positive rate is defined as
the fraction of samples the attack labels as ``member'' but in fact were
not in the training dataset; and conversely the false negative rate is
the fraction of samples labeled as ``nonmember'' but in fact \emph{are}
in the training dataset.

It is possible to (provably) prevent these forms of attacks by 
constructing an algorithm that is \emph{differentially private}.
%
%
Kairouz \emph{et al.}~\cite{kairouz2015composition} prove that for any $(\varepsilon,\delta)$-DP algorithm,
we can bound the ratio between the
true positive rate (TPR) and false positive rate (FPR) by
\begin{equation}
   {TPR - \delta \over FPR} < e^{\varepsilon}.
   \label{eq:tpr-fpr}
\end{equation}

This bound makes training models with DP desirable as it
prevents even the weakest forms of privacy attacks.
While there are many techniques available to train models with
differential privacy [cites], these methods often sacrifice model accuracy
in order to achieve provable privacy.

\paragraph{Backpropagation Clipping} is a differentially private
training algorithm to train deep neural networks.
It modifies the standard DP-SGD algorithm which we now briefly review.

In standard stochastic gradient descent, a model's parameters are updated
by taking the gradient (with respect to the parameters) of the loss
of the model on a particular set of training examples $B \subset D$,
formally, $\theta_{i+1} = \theta_{i} - \nabla_{\theta_{i}} \mathcal{L}(f_{\theta_{i}}, B)$.
This process slowly learns a set of parameters $\theta$ that reach
(approximately, usually, and without guarantee) minimum training loss on the training set $D$.

DP-SGD modifies this process by first \emph{clipping} each example's gradient
to have norm at most $C$,
to bound its \emph{sensitivity} (the maximum impact of a single sample), and then adding noise
$n \sim \mathcal{N}(0, \sigma^2)$ sampled from a Normal distribution.
Formally,
$\theta_{i+1} = \theta_{i} - \left(\sum_{x \in B}\text{clip}_C\big(\nabla_{\theta_{i}} \mathcal{L}(f_{\theta_{i}}, x)\big)\right) + n$.
Training in this way allows one to produce a proof that the algorithm
is $(\varepsilon, \delta)$ differentially private.

Backpropagation Clipping alters this process slightly.
Instead of clipping each example's final gradient, clipping is performed per-layer, during the model's forward pass and backward pass, and composition is applied per layer. Specifically, the input to each layer is clipped to norm $C_1$ (on a per-example basis), and the layer's backward signal is clipped to norm $C_2$ (also on a per-example basis). This dual clipping ensures that each example's gradient, for that layer, has norm bounded by $C_1\cdot C_2$. Gaussian noise proportional to this sensitivity is then added to each layer's gradient.

\paragraph{Auditing machine learning models.}
Prior work has shown that it is possible to \emph{audit} a differentially
private machine learning pipelines to establish a lower bound on the
privacy parameter $\varepsilon$ (as opposed to the proofs that give an upper bound).
Because any $(\varepsilon,\delta)$-differentially private model
cannot be vulnerable to a membership inference attack with 
a TPR-FPR ratio above roughly $\exp(\varepsilon)$, we can establish a \emph{lower bound}
on $\varepsilon$ by developing a strong membership inference
attack.
By computing the attack's TPR and FPR, we may compute this $\varepsilon$ using Equation~\ref{eq:tpr-fpr}.

In this paper we will use auditing instead to show an algorithm is flawed,
by demonstrating the (empirical) lower bound is \emph{larger} than the (``proved'')
lower bound.
As discussed in \cite{jagielski2020auditing, nasr2021adversary}, 
we must be careful about the statistical validity of the results.
For example, if our attack
happened to identify just one sample (out of $N$) as a true positive
and all others it predicted negative, then the TPR would
be $1/N$, with a FPR of $0$.
Then as long as $1/N > \delta$, there is no finite lower bound and $\varepsilon=\infty$! However, this is not statistically significant.
After just two more trials, 
we could possibly observe a false positive and a false negative, giving a true positive rate and false positive rate both of $1/(N+1)$, which cannot give any lower bound of $\varepsilon>0$. %
To fix this, auditing analysis techniques~\cite{jagielski2020auditing, nasr2021adversary} use Clopper-Pearson 
confidence intervals~\cite{clopper1934use} to establish a probabilistic lower bound on $\varepsilon$, by lower bounding TPR and upper bounding FPR.

\section{Auditing Backpropagation Clipping}

We now show how auditing can check the correctness of one recent training scheme:
Backpropagation Clipping.
We find that
the official implementation~\cite{stevens2022backpropagation} 
does not give the $(\varepsilon,\delta)$-differential privacy
guarantee that is claimed.

Here, we assume Backpropagation Clipping operates as a black-box,
receiving as input a training dataset $D$ and returning as output a trained machine learning model $f_\theta \gets \mathcal{T}(D)$.

We audit the privacy of this algorithm by constructing a pair of datasets
$D,D'$ that differ in one example, but where it is possible to distinguish between
a model trained on $D$ from a model trained on $D'$. This gives a membership inference attack on the sample, 
with a TPR/FPR ratio statistically significantly $(p \ll 10^{-10})$
higher than should be possible if the model achieved
$(0.21,10^{-5})$-DP.

While differential privacy guarantees that the distinguishing game
should fail for \emph{all} adjacent dataset $D,D'$, we will show that the
distinguishing attack actually succeeds even when $D$ is the MNIST training set.
This means that Backpropagation Clipping is not only insecure in some
hypothetical and pathological worst-case setting. In real situations
that could be encountered when training on standard data, the scheme does not provide the promised privacy guarantees.

\paragraph{Experimental setup.}
We perform our analysis on the MNIST dataset of $60{,}000$ hand-written digits
from 0 to 9.
We add to MNIST a single poisoned sample $(x_p, y_p)$ in order to
get a $60{,}001$-example dataset that we call MNIST'
(we discuss later how we construct this poisoned example).

We run Backpropagation Clipping with the hyperparameters that give the most
accurate MNIST model, listed in Figure 2 of Stevens \emph{et al.}~\cite{stevens2022backpropagation}:
\begin{itemize}
\setlength\itemsep{-.2em}
\item Epochs: 25
\item Batch size: 4096
\item Input clipping norm: 1.0
\item Gradient clipping norm: 0.01
\item Per iteration rho: $10^{-5}$
\end{itemize}
When run with the official implementation\footnote{\url{https://github.com/uvm-plaid/backpropagation-clipping}}, these parameters return an MNIST
model that reaches $98.9\% \pm 0.1\%$ accuracy with a privacy guarantee of
$(0.21, 10^{-5})$-DP.
These parameters completely specify the training algorithm $\mathcal{T}$.

Our membership inference $\mathcal{A}$ is a simple loss-based membership
inference attack:
to predict whether or not an example $(x, y)$ is contained in the models
training dataset, we carefully choose a threshold $\tau=2.64$ (the method to choose this threshold is again discussed later) and
report ``member'' if $\mathcal{L}(f_\theta, x, y) < \tau$, or
otherwise ''nonmember''.
%

\paragraph{Our auditing analysis.}
We are able to empirically refute the claimed DP guarantees ($p \ll 10^{-10}$).
To do this, we train $100{,}000$ models with Backpropagation Clipping on
MNIST and another $100{,}000$ on MNIST'.
Among the models trained on MNIST, 
there are $174$ false positives where the loss is less
than the threshold, $\mathcal{L}(f, x_p, y_p) < \tau$.
And for the models trained on MNIST', there are $4{,}922$ true positives (again, $\mathcal{L}(f', x_p, y_p) < \tau$).
Using standard Clopper-Pearson confidence intervals for binomial proportions, we find that the false positive rate is almost
certainly 
less than $274/10^5$, and 
the true positive rate is almost certainly more than $4491/10^5$, at a joint p-value of $p<10^{-10}$.
Therefore, by Equation~1, and assuming a value of $\delta=10^{-5}$,
we can say with near certainty that $\varepsilon > 2.79$.
This refutes the claim that the algorithm is $(0.21, 10^{-5})$-DP,
and in fact shows that the lowest possible value of $\varepsilon$
is at least $10\times$ higher than has been claimed.

As a note, even though our analysis trained an absurd number of models\footnote{In total we trained over $250,000$ models on 8 V100 GPUs for 50 hours.},
just $1{,}000$ would have sufficed to reject the claimed
$\varepsilon=0.21$ with $99\%$ confidence.
This is something that could reasonably be done without much effort:
because these MNIST models train at a rate of two a minute on a
V100 GPU, this would take roughly 8 GPU hours to complete.
We perform the additional experiments only to establish an upper bound $10\times$ higher than has been ``proven'', and to obtain a figure that
looks nicer.

\subsection{Implementation Details}

The above attack defined (without motivation) both a
poisoned sample $(x_p, y_p)$
and a membership inference attack $\mathcal{A}$.
Below we describe the (heuristic) strategy we
used to choose these.

\paragraph{Choosing the poisoned sample.}
Our attack requires an example $(x_p, y_p)$ that we will insert into
the MNIST training set.
For simplicity, as mentioned above, we choose an image from the MNIST
test set, mislabel it, and insert the mislabeled image into the training set.
(Prior work has found using adversarial examples is even better.
We found mislabeling to be sufficient.)
To choose which MNIST test image we should use as a poisoned
sample, we perform a preliminary experiment where we run our attack, with $1{,}000$ models, for each the first $25$ images in the MNIST test set. This amounts to training $26{,}000$ models total ($1{,}000$ for each sample, and $1{,}000$ with the original training set).
We then select the image from this set where our attack achieves the largest $\varepsilon$ lower bound. 
Following the backdoor sample intuition from \cite{jagielski2020auditing}, we insert a checkerboard pattern in the corner of the image to hopefully make it more distinguishable and provide a larger $\varepsilon$~\footnote{Note that our strategy to construct this poisoned sample has 
\emph{no} theoretical justification, but is reasonable and simple to implement.
Prior work on auditing has proposed strategies for audit sample selection~\cite{jagielski2020auditing, nasr2021adversary} - we expect these samples would have performed similarly or better than ours, and future work may be able to design even stronger principled audit samples.}.
Note that, had this specific sample failed to refute privacy, differential privacy still makes a guarantee about \emph{all} such samples, so the failure would not have confirmed the model's privacy.
In order to ensure statistical validity of our results,
we throw away all models we train for this initial experiment and train
new models from scratch for all other experiments.

\paragraph{Choosing the threshold $\tau$.}
To get a successful membership inference attack, we need to identify the best possible loss threshold $\tau$, where examples with loss less than $\tau$
are predicted as members.
As is done in prior work on auditing, we perform an initial run of auditing with the sole purpose of identifying the best threshold, which we will use to distinguish future models.
We train $2{,}000$ models with the Backpropagation Clipping algorithm;
half of these models are trained on MNIST and the other half on MNIST$'$
using the best poisoned sample identified above.
For each trained model, we then record the model's loss $\mathcal{L}(f, (x_p, y_p))$
on the example $(x_p, y_p)$.
This gives us a distribution of loss values when we train on the poisoned sample,
and when we do not.
Using these samples, we sweep over all possible thresholds $\tau$ to identify which value gives the largest TPR/FPR ratio.
We find the best value occurs at $\tau=2.64$---this is visualized in Figure~1.
After identifying this threshold, we discard the $1{,}000$ models for this experiment again to ensure statistical validity.

\section{Debugging The Privacy Leak}

We have demonstrated that the implementation of Backpropagation Clipping does not provide the claimed level of differential privacy. We thus set out to understand what caused this discrepancy.

We found that the issue was an implementation error, where the sensitivity of each layer's gradient was mistakenly reduced by a factor of the batch size $|B|$. As a result, the noise added to the gradients was also too small by a factor $|B|$.

This error might appear to be a simple mistake in translating the paper's algorithm to code. Yet, very similar mistakes have previously been made in other differential privacy implementations. 
(We are aware of two prior examples: \cite{Park_VIPS_fix} which was identified and fixed by the authors prior to peer-reviewed publication, and \cite{tfprivacyissue} which is currently ongoing.) 
We thus describe this bug in more detail below, along with some suggestions on how to detect and mitigate similar bugs in the future.

\paragraph{Calibrating sensitivity to batch size.}

The implementation of Backpropagation Clipping uses the standard cross-entropy loss, which averages the losses of each example in a batch:
\[
\mathcal{L}(f, B) = \frac{1}{|B|}\sum_{(x, y) \in B} \mathcal{L}(f, x, y)
\]
Thus, the gradient of the batch loss is the average of per-example gradients in the batch:
\[
\nabla_\theta \mathcal{L}(f, B) = \frac{1}{|B|}\sum_{(x, y) \in B} \nabla_\theta \mathcal{L}(f, x, y)
\]

If each per-example gradient $\nabla_\theta \mathcal{L}(f, x, y)$ was guaranteed to have norm bounded by $C_1\cdot C_2$, then the sensitivity of the batch gradient would be $\frac{C_1 \cdot C_2}{|B|}$. This is exactly what is defined in the authors' implementation. What then is wrong here?

The mistake comes from the assumption that it is the per-example gradient 
that is being clipped, i.e, $\| \nabla_\theta \mathcal{L}(f, x, y) \| \leq C_1 \cdot C_2$, when it is actually the per-example gradient \emph{divided by the batch-size}: $\| \frac{1}{|B|}\nabla_\theta \mathcal{L}(f, x, y) \| \leq C_1 \cdot C_2$. Thus, the sensitivity of the batch gradient should be  $C_1 \cdot C_2$, a factor $|B|$ larger than defined in the code.

To see this, each example's error that is backpropagated through the network is scaled by the partial derivative:
\[
\frac{\partial \mathcal{L}(f, B)}{\partial \mathcal{L}(f, x, y)} = \frac{1}{|B|}\;.
\]
The implementation ensures that the backpropagated error signal of each example is clipped to norm $C_2$ (and \textbf{not} to norm $\frac{C_2}{|B|}$).
The batch size is thus already implicitly accounted for in the Backpropagation Clipping, and should not appear in the sensitivity calculation as well.

\paragraph{Recommendations.}

A simple sanity check on the gradient sensitivity can be computed using the triangle inequality. If the sensitivity of the batch gradient $g$ is claimed to be $s$, and the batch size is $B$, then it must hold that $\| g \| \leq |B| \cdot s$. We find that this test is violated in the original implementation (where $s = \frac{C_1 \cdot C_2}{|B|}$). If we instead define $s=C_1 \cdot C_2$, this sanity check passes (but the model utility is degraded significantly due to the larger noise addition).

We further recommend that papers that propose new differentially private algorithms provide formal algorithm descriptions that match their intended implementation as closely as possible. For example, if an algorithm clips certain quantities at a per-example level, it is useful for the formal description of the algorithm to make individual examples explicit (see~\cite{abadi2016deep} for an example).

\section{Conclusion}

Unlike other areas of secure or private machine learning
where results are often empirical observations that appear true
without formal justification
(e.g., defenses to adversarial examples),
the appeal of differential privacy is that it gives
\emph{provably correct} results.
Unfortunately, as we have seen here,
while producing correct proofs is a necessary prerequisite to training private machine learning models, it is important to also get all the subtleties right.

The other lesson from this is that future papers should follow the direction of backpropagation clipping~\cite{stevens2022backpropagation} and
release code along with algorithms. 
The implementation provided by the authors
faithfully reproduces every aspect of the paper and without this code, our
analysis would have been significantly more complicated.

We encourage future work to use strong auditing
techniques even when the results are provably correct.

\bibliographystyle{alpha}
\bibliography{paper}

\end{document}